\documentclass[conference]{IEEEtran}
\IEEEoverridecommandlockouts
\newcommand{\cmark}{\ding{51}} 
\newcommand{\xmark}{\ding{55}} 
\usepackage{booktabs}
\usepackage{graphicx}
\usepackage{cite}
\usepackage{amsmath,amssymb,amsfonts}
\usepackage{algorithmic}
\usepackage{graphicx}
\usepackage{textcomp}
\usepackage{xcolor}
\usepackage{pifont}
\usepackage{url}
\def\BibTeX{{\rm B\kern-.05em{\sc i\kern-.025em b}\kern-.08em
    T\kern-.1667em\lower.7ex\hbox{E}\kern-.125emX}}
\begin{document}

\title{Multi-Task Visual Perception Network with LLM Conditioning for Autonomous Navigation\\
}
\author{
	\IEEEauthorblockN{Praveen Kumar, K.R. Guruprasad\textsuperscript{\dag}, and Tushar Sandhan}
	\IEEEauthorblockA{
		Department of Electrical Engineering, \textsuperscript{\dag}Department of Mechanical Engineering\\
		Indian Institute of Technology Kanpur, India\\
		\{praveenk20, krgprao, sandhan\}@iitk.ac.in
	}
}

\maketitle

\begin{abstract}
Long-term navigation for service robots faces 
critical challenges like the accumulation of odometry drift and
sensor error, which progressively degrade 2D
maps and renders traditional path planning algorithms (e.g., A*,
RRT*, DiPPer, ViT-A*) ineffective over time. To
address this, we propose a user-friendly, interactive framework
that eliminates the reliance on globally consistent maps. Our
approach integrates visual perception with Large Language
Models (LLM) to interpret user commands via text or voice. Instead
of relying on a drift-prone global map, the system generates a
sequential action plan based on local visual cues and egocentric
geometric instructions. These action plans are executed sequentially,
allowing the robot to navigate known and unknown environments
safely. By resetting localization relative to immediate targets, our
framework effectively works with a minimum accumulation drift strategy,
ensuring accurate, efficient, and collision-free navigation without
the maintenance overhead of traditional mapping. Experiments on real-world and simulated data have shown significant improvements over other methods. Our source code is publicly accessible at \url{https://github.com/PraveenSingh24/VL-Navigation}.
\end{abstract}


\section{Introduction}
Service robots are increasingly deployed in dynamic indoor settings, such as residences and offices, where they must operate autonomously for extended periods. However, these environments are characterized by narrow passages and complex visual clutter, which often degrade traditional map-based navigation systems. 

We propose a paradigm shift towards human-centric navigation, inspired by how humans verbally coordinate tasks. Consider an example scenario where a robot's admin commands a robot to ``fetch a water bottle from the kitchen," or in a navigation context, a user might guide the robot using egocentric geometric and visual mark instructions such as: ``Move 5 meters forward, take a right turn, and stop near the statue (Elephant)." Robots use natural language to identify local visual landmarks and geometric subgoals during a mission. It resets their navigation at each stage, significantly reducing long-term drift accumulation.
Path planning has evolved from classical search-based algorithms (A*~\cite{4082128}, D* Lite~\cite{koenig2002d}) and sampling-based techniques (RRT~\cite{lavalle2001randomized}, PRM~\cite{kavraki2002probabilistic}) to modern hybrid architectures (Informed RRT*~\cite{gammell2014informed}, Hybrid A*~\cite{dolgov2008practical}) and data-driven neural planners (ViT-A*~\cite{liu2023vit}, DiPPer~\cite{liu2024dipper}). While these modern approaches offer improved efficiency or adaptability, they remain constrained by a common bottleneck: the requirement for a static, globally accurate map. 
\begin{figure}[t]
	\centering
	\includegraphics[width=0.8\linewidth, trim={9.0cm 5.0cm 9.0cm 3.0cm}, clip]{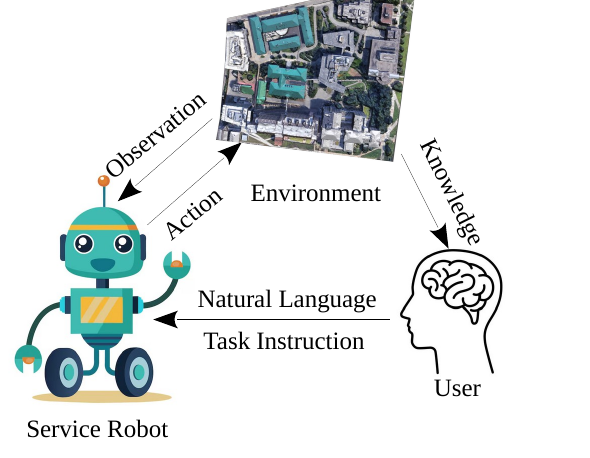}
	\caption{Overview of the proposed scenario. The user leverages prior spatial knowledge of the environment to provide natural language task instructions to the service robot. The robot interprets these instructions and interacts with the environment through continuous observation and action, executing navigation tasks sequentially without relying on a global map.}
	\label{fig:intro}
\end{figure}
Reliance on globally consistent mapping constitutes a primary cause of failure for continuous service operations in
an open, dynamic, and unstructured home environment~\cite{zhang2025environment}. Traditional SLAM frameworks assume a static world, which degrades rapidly in dynamic human-centric environments where furniture and layout change frequently~\cite{biber2005dynamic}. Further, the pneumatic tire slip and skid-steering produce non-linear odometry errors that state-of-the-art SLAM algorithms cannot entirely suppress over long-duration deployments~\cite{zhou2024visual}. In visually complex, narrow passageways, the odometry drift can lead to collision or mission failure. Consequently, the global map differs from real-world settings, diminishing the ability to localize and reach goals and increasing the need for mapless, egocentric navigation strategies~\cite{lin2021reinforcement}. Our work does not use a static (Known) and an online-explored (Unknown) global map for path finding, as in traditional methods and latest learning models do; instead, it uses them to demonstrate our system's performance. Additionally, these systems demand precise metric goal coordinates $(x,y)$, creating a usability gap; human users naturally communicate tasks via semantic instructions (e.g., ``go to the kitchen") rather than providing global coordinates on a distorted digital map.
\begin{figure*}[t]
	\centering
	\includegraphics[width=0.8\linewidth, trim={1.0cm 3.0cm 1.0cm 2.8cm}, clip]{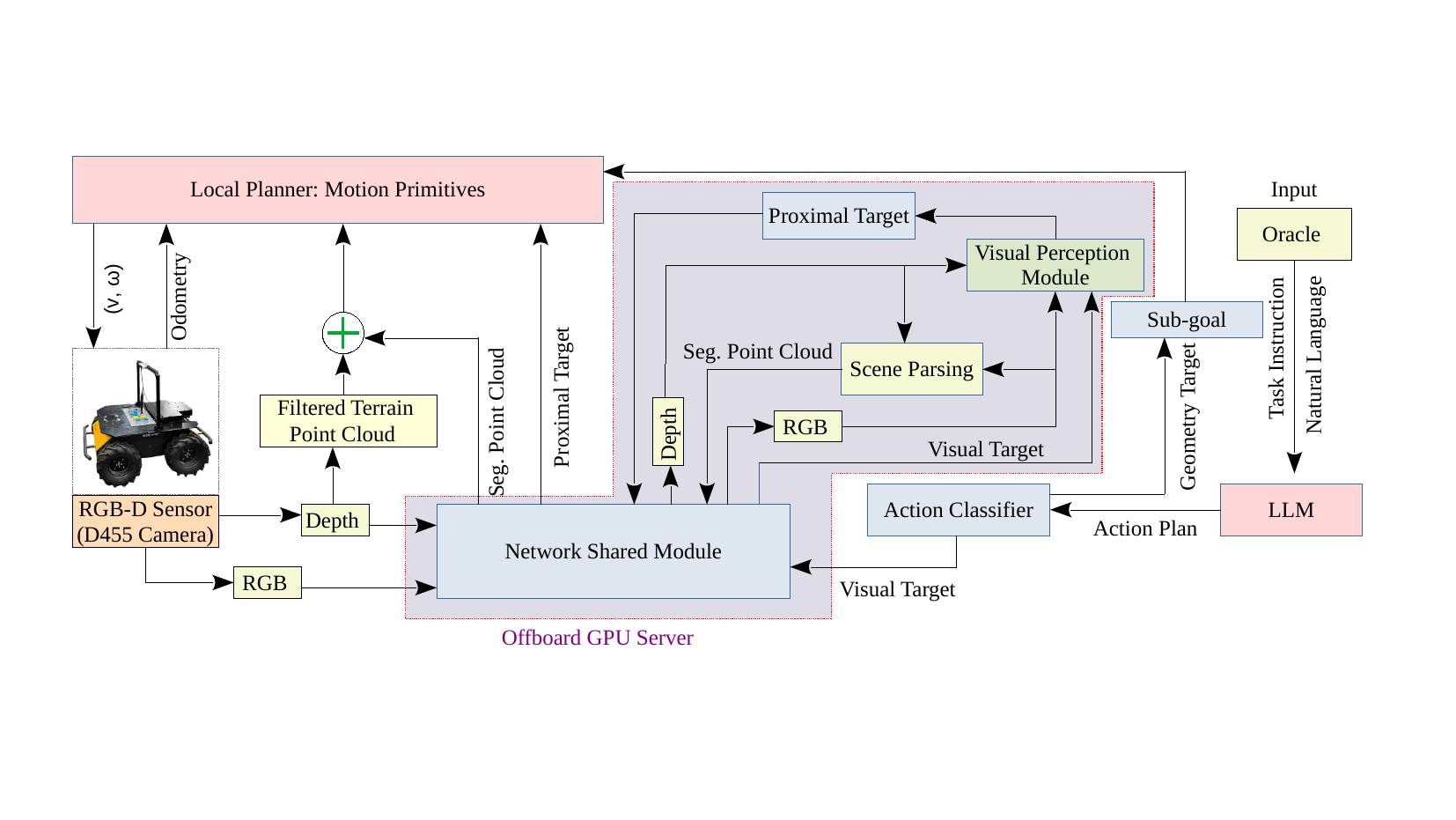}
	\caption{Overview of the proposed vision-language navigation framework, which is distributed across two nodes: the onboard client (Husky A200) and an offboard GPU server for heavy computation. The workflow begins with Oracle Task Instructions (Natural Language), which are processed by LLM to generate an action plan. The Action Classifier parses this plan into geometric or visual sub-goals. To enable safe navigation, the Scene Parsing module generates a Segmented (Seg.) Point Cloud—representing specific semantic regions—which is concatenated with the filtered terrain point cloud. Finally, the Motion Primitive Local Planner computes optimal velocity commands $(\upsilon,\omega)$ to drive the robot toward the Proximal Target.}
	\label{bdiagram}
\end{figure*}
As discussed in the recent survey on tasks and methods in vision-and-language navigation~\cite{gu2022vision}, the field is shifting towards agents that can seamlessly bridge the gap between visual perception and natural language communication. Recent works such as DyNaVLM~\cite{ji2025dynavlm} and VLingNav~\cite{wang2026vlingnav} investigate vision-language navigation using learned memory and reasoning mechanisms, with VLingNav further introducing an adaptive chain-of-thought (AdaCoT) strategy for dynamic decision-making. AdaVLN~\cite{loh2024adavln}, in contrast, provides a simulator and benchmark for evaluating such approaches. At the same time, these approaches are effective in controlled environments. These methods often rely on computationally intensive representations or extensive training, limiting their robustness and deployability in real-world settings. We align with this direction by introducing a framework in which the robot acts as an intelligent assistant, processing natural-language user commands and combining current and past visual context.

Rather than planning a long path on a noisy and drifted map, we propose the LLM decomposes the user's intent into a sequence of action tasks. It enables the service robot to navigate to a semantic goal (e.g., a specific object or room) accurately and efficiently by grounding navigation in local visual observations rather than global coordinates. Our framework can be deployed to provide continuous uninterrupted services in a critical, high-risk mission. Even in a narrow passage setup, our framework worked well. 

\section{Proposed Method}
This section details the proposed vision-geometric navigation framework designed for wheeled service robots operating in human-centric environments. To ensure accessibility for non-expert users in daily scenarios, the system interacts via natural language, allowing operators to issue intuitive task instructions without requiring technical expertise. The framework intelligently interprets these guidelines to navigate workspaces containing critical visual constraints, effectively filling the gap between human intent and robotic execution. As shown in Fig. 2, the architecture integrates an LLM (Gemini Pro), a Visual Perception Module, and a Motion Primitive Local Planner to execute these instructions while mitigating odometry drift in constrained spaces.

\subsection{Task Instruction Processing Module}
For service robots to evolve from industrial tools to ubiquitous assistants in homes and offices, they must be accessible to users unfamiliar with technical robotics.
For instance, the system parses:

\begin{itemize}
	\item {Object Retrieval:} ``Bring a water bottle from the kitchen.''
	\item {Hybrid Geometric \& Semantic Navigation:} ``Go ahead 10m, turn right at the Lotus statue, and proceed to the pharmacy.''
	\item {Social \& Contextual Interaction:} ``Hi Buddy, please deliver this file to Mr. David and report back to me.''
\end{itemize}
\begin{figure}[t]
	\centering
	\includegraphics[width=1.0\linewidth, trim={6.0cm 5.0cm 5.0cm 4.5cm}, clip]{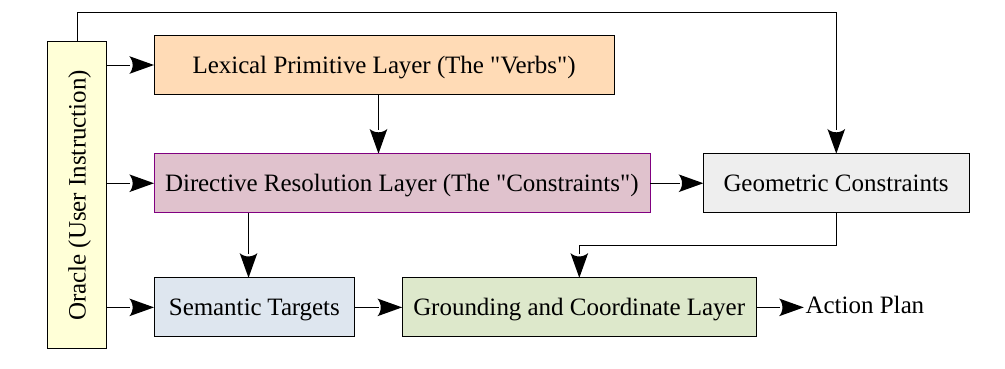}
	\caption{The multi-layered architecture for converting user task instructions into robot action plan. The system resolves lexical ambiguity and distinguishes between geometric goal and semantic target constraints before grounding them into a unified robot system for execution.}
	\label{llmw}
\end{figure}
We propose a novel Hierarchical Task Network (HTN) architecture as illustrated in Fig.~\ref{llmw}, designed for translating natural language instructions into robot actions in service and industrial robots. The process begins at the Lexical Primitive (e.g., FIND, MOVE) Layer (Level-1), where core action verbs are extracted to identify task intent. This is followed by the Directive Resolution Layer (Level-2), which bifurcates the instruction into dual pathways: an egocentric branch for geometric constraints relative to the robot's local frame, and an allocentric branch for semantic targets within the global coordinate frame. Finally, the Grounding and Coordinate Layer (Level-3) parameterizes these directives into precise spatial data and dispatches them to specialized ROS execution nodes via a unified, serialized action plan.
\begin{figure}[t]
	\centering
	\includegraphics[width=1.0\linewidth, trim={6.0cm 5.0cm 6.0cm 4.5cm}, clip]{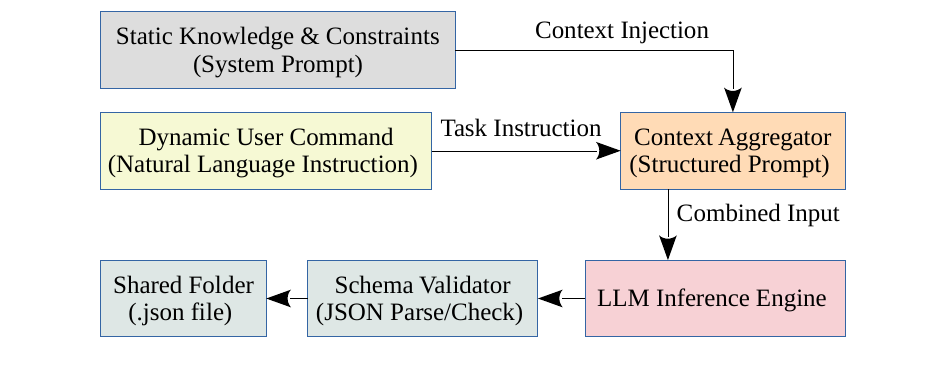}
	\caption{Structured prompt is created via injecting static system constraints into the context window alongside the user task instruction. LLM translates the natural language request into a machine-readable format.}
	\label{llm}
\end{figure}
To ensure reliability and safety, we use Gemini Pro for its low-latency multimodal reasoning; hallucinations and undefined behaviors are mitigated by constraining outputs to a closed action set and enforcing a schema validator (Fig.~\ref{llm}). The architecture illustrated in Fig.~\ref{llm} adopts prompt engineering via our action graphs to bridge the gap between unstructured natural-language commands and deterministic robotic control. A composite prompt is constructed by combining the user’s instruction with a set of domain-specific constraints, including available object categories and predefined action primitives.

The resulting output is therefore not free-form text but a deterministic symbolic representation that the robot’s control pipeline can directly interpret.
The Action Classifier is implemented as a rule-based semantic parser that transforms natural language task instructions into an ordered sequence of executable commands. When a visual subgoal cannot be found, the robot initiates a bounded exploration within a predefined obstacle-free region to re-detect the next goal. For an unsuccessful attempt, the system queries the user with contextual feedback and regenerates the action plan accordingly. 
\subsection{Action Plan Decoding and Subgoal Estimation}
The action plan encodes high-level navigation commands in a structured format suitable for robotic execution. We classify these commands into two categories: perceptual targets, which require visual reasoning, and egocentric geometric targets, which define motion primitives relative to the robot's current pose.

Egocentric geometric targets are computed directly from the instruction semantics through context parsing and are translated into spatial subgoals expressed in the robot's egocentric coordinate frame. These subgoals guide the navigation controller toward the desired waypoint in a step-wise manner.
Let the robot's pose in the global coordinate frame be defined as $\mathbf{R}_r = [x_r, y_r, \theta_r]^T$, where $(x_r, y_r)$ denotes the position and $\theta_r$ represents the heading. Correspondingly, the egocentric motion command extracted from the instruction is represented as $\Delta \mathbf{p} = [\Delta x, \Delta y]^T$, where $\Delta x$ and $\Delta y$ define the required displacement within the robot's local coordinate frame.

The global target position $\mathbf{G}_c = (x_g, y_g)$ is computed by transforming the egocentric motion into the global frame using the relation $\mathbf{G}_c = [x_r, y_r]^\top + \mathbf{R}(\theta_r) [\Delta x, \Delta y]^\top$, where the rotation matrix $\mathbf{R}(\theta_r) = \left[ \begin{smallmatrix} \cos\theta_r & -\sin\theta_r \\ \sin\theta_r & \cos\theta_r \end{smallmatrix} \right]$ aligns the local displacement with the robot's current heading $\theta_r$ relative to the global origin.
This computed global coordinate $\mathbf{G}_c$ is then published as a waypoint to a modified version of the local planner described in~\cite{cao2022autonomous}, which we adapt to our navigation setup and task objectives. This allows the robot to execute the relative motion specified by the natural language instruction.

Visual perception targets are categorized into two fundamental perception tasks: (i) object detection and (ii) person re-identification (ReID). The selection of the appropriate perception module is automatically determined based on the action keyword extracted from the generated action plan. For object-centric goals, the framework employs a YOLOv8-based detector, while person-specific tasks are handled using an OSNet-based ReID model~\cite{zhou2019omni} trained on our custom dataset. During execution, RGB images acquired from the Intel RealSense D455 camera are processed online by the selected perception network to localize the target and generate a bounding box, which is subsequently used for downstream spatial reasoning and navigation.
Let the bounding box be defined as $\mathcal{B} = (u_{\min}, v_{\min}, u_{\max}, v_{\max})$, and its center pixel be $(u_c, v_c)$. A robust depth estimate is obtained by computing the median depth value within the bounding box region, filtered to remove invalid or noisy measurements
\begin{equation}
Z = \operatorname{median}\left\{ D(u,v) \mid (u,v) \in \mathcal{B},\; Z_{\min} < D(u,v) < Z_{\max} \right\}.
\end{equation}
Using the camera intrinsic matrix $\mathbf{K}$, the 3D coordinates of the detected object in the camera optical frame are recovered as $X_c = (u_c - c_x) Z / f_x$, $Y_c = (v_c - c_y) Z / f_y$, and $Z_c = Z$.
Following the standard ROS coordinate convention, these coordinates are transformed into the robot base frame as $X_r = Z_c$, $Y_r = -X_c$, and $Z_r = 0$. The resulting point is then transformed into the global coordinate frame using the robot’s pose and the corresponding rigid-body transformation $[x_m, y_m]^\top = \mathbf{R}(\theta_r) [X_r, Y_r]^\top + \mathbf{t}$, where $\mathbf{R}(\theta_r)$ and $\mathbf{t} = [x_r, y_r]^T$ represent the rotation and translation from the robot frame to the map frame. To ensure safe navigation, the robot does not directly approach the detected object but instead computes a subgoal at a fixed stopping distance $d_s$ from the target. Let $(x_r, y_r)$ denote the robot position and $(x_m, y_m)$ the target position in the map frame. The final navigation goal is computed by first determining the Euclidean distance $d = \sqrt{(x_m - x_r)^2 + (y_m - y_r)^2}$ and bearing $\theta = \tan^{-1}((y_m - y_r)/(x_m - x_r))$ to the target. Subsequently, the goal coordinates, adjusted for a safe stopping distance $d_s$, are derived as $x_g = x_r + (d - d_s)\cos\theta$ and $y_g = y_r + (d - d_s)\sin\theta$.
This target $(x_g, y_g)$ is then published as a navigation waypoint, allowing the robot to approach the detected object while maintaining a safe and controlled distance.
\subsection{Context-Aware Environmental Perception}
The robot perceives and interprets its surroundings to support safe, task-aware navigation. The perception module enables the robot to identify semantically meaningful elements such as color-marked restricted zones, electrical cables lying on the ground, objects placed along the navigation path (e.g., mobile phones or documents), and chemically contaminated or hazardous surface regions. The relevance of these elements may vary depending on the operational context, environmental conditions, and user-defined task priorities.

\begin{figure}[t]
	\centering
	\includegraphics[width=1.0\linewidth, trim={5.0cm 5.0cm 5.0cm 5.5cm}, clip]{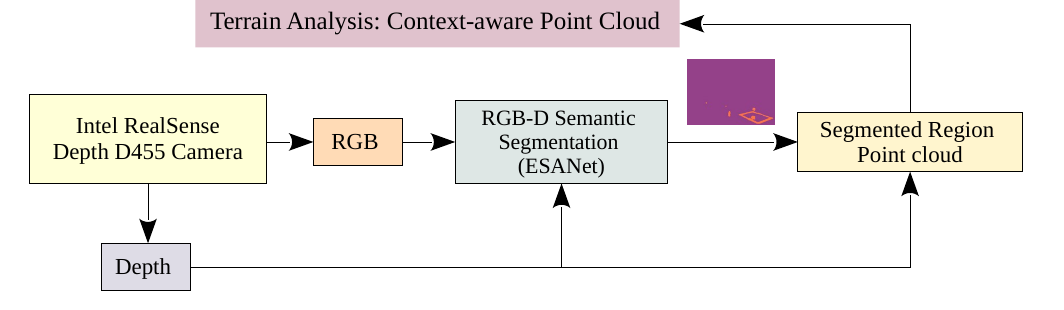}
	\caption{Visual perception to point cloud generation pipeline. The RGB–D stream acquired from the Intel RealSense D455 camera is processed by an RGB-D semantic segmentation network (ESANet) to extract task-relevant regions. The resulting segmented mask is projected into 3D space using depth information to generate a semantic point cloud, which is subsequently utilized for terrain analysis and context-aware navigation.}
	\label{contextAware}
\end{figure}
 
\subsection{Motion Primitive Generation and Selection}

To ensure safe and efficient autonomous navigation in complex environments, the local planner evaluates a discrete set of candidate trajectories against semantic perception data and geometric constraints. This process is divided into two stages: the retrieval and transformation of pre-computed motion primitives, and the optimal selection based on a context-aware cost function.

\subsubsection{Pre-computed Path Library and Transformation}
\begin{figure}[t]
	\centering
	\includegraphics[width=1.0\linewidth, trim={4.0cm 3.0cm 4.0cm 3.5cm}, clip]{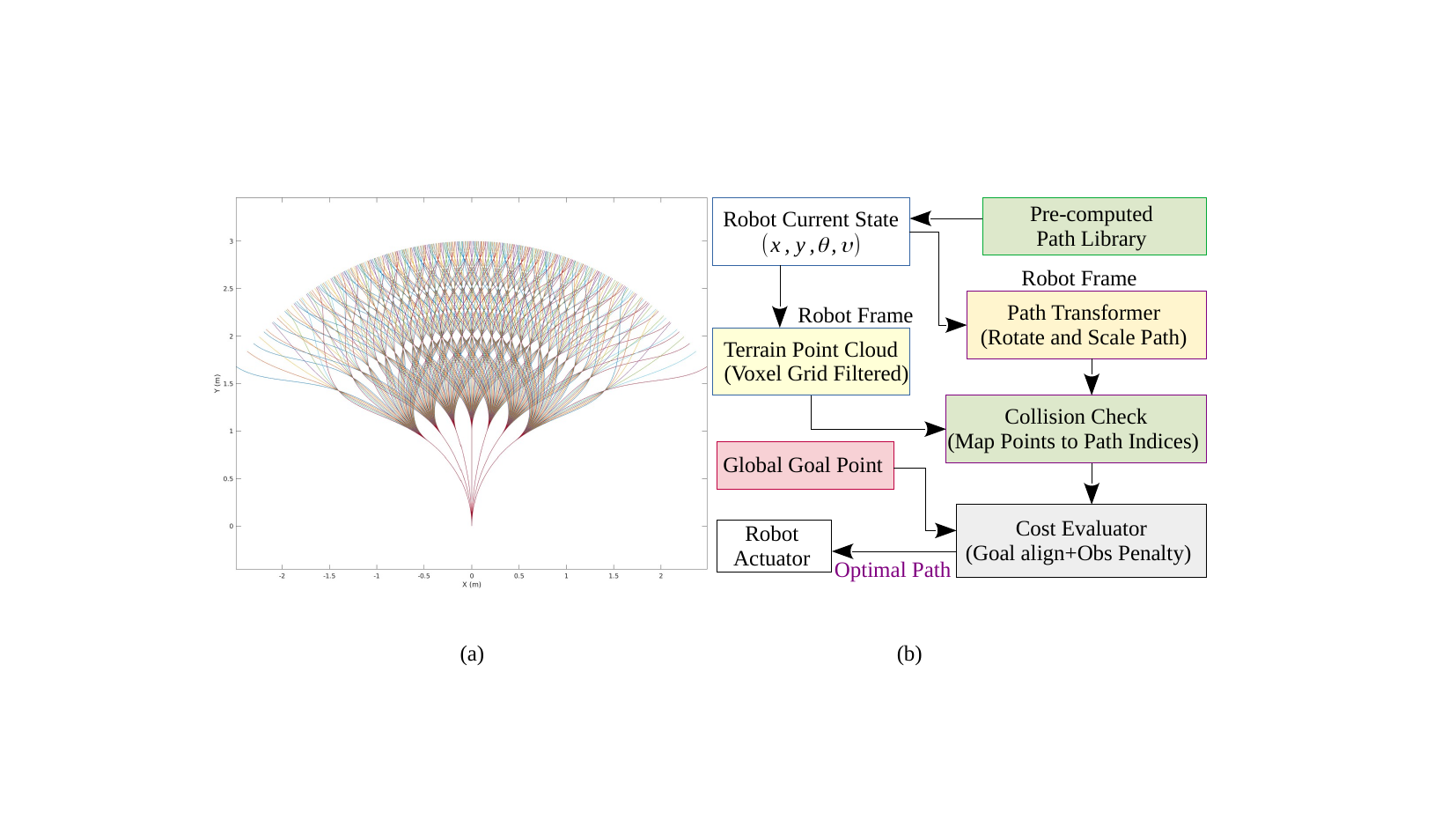}
	\caption{(a) Set of candidate motion trajectories produced using the motion primitive framework. These trajectories represent feasible forward motions under kinematic constraints and are evaluated against environmental and semantic constraints to select an optimal navigation path. (b) Data flow of the local planner. The process begins by synchronizing the pre-computed path library and sensor data to the robot's current state. Feasible paths are filtered through a voxel-grid collision check, and the final trajectory is chosen by minimizing a cost function to ensure safe and efficient navigation.}
	\label{localPlanner}
\end{figure}

Instead of solving the boundary value problem (BVP) for real-time trajectory generation, which can be computationally expensive, we use a library of precomputed motion primitives. As shown in Fig.~\ref{localPlanner}~(a), this library consists of 343 kinematically feasible trajectories generated offline. These primitives represent a diverse set of maneuvers that respect the mobile robot's non-holonomic constraints.

During runtime, the planner adapts these static templates to the robot's dynamic state $\mathbf{x}_t = [x, y, \theta, v]^T$. The selected primitives are transformed from the library frame to the robot's local coordinate frame using a rigid-body transformation and scaled according to the current vehicle velocity $v_{curr}$. It ensures that the search space expands faster to look further ahead and contracts more slowly for precise maneuvering.
\subsubsection{Feasibility and Collision Checking}
The transformed candidates are evaluated for feasibility using a temporary local representation—an egocentric voxel grid constructed from real-time depth frame data and ESANet~\cite{seichter2021efficient}. Unlike persistent global SLAM maps used for long-term planning, this transient representation is employed solely for immediate collision-free trajectory selection. A lookup table maps each voxel index to the subset of motion primitives that intersect it. If a primitive traverses an occupied voxel—defined as a point where the terrain height or semantic class indicates an untraversable obstacle—it is flagged as invalid and removed from the candidate set.
\subsubsection{Optimal Path Selection}
The optimal trajectory $\tau^*$ is selected from the set of candidate motion primitives by minimizing a composite cost function $J(\tau)$. The scoring mechanism balances goal alignment with safety from both physical and semantic hazards,

\begin{equation}
J(\tau) = w_g \cdot J_{goal}(\tau) + w_o \cdot J_{obs}(\tau) + w_s \cdot J_{semantic}(\tau)
\end{equation}

	where $J_{goal}(\tau)$ represents the heading alignment error, quantifying the deviation between the robot's orientation at the end of the primitive ($x_{end}, y_{end}, \theta_{end}$) and the bearing to the subgoal $\mathbf{G}_c$ and $\operatorname{atan2}(\cdot, \cdot)$ denotes the four-quadrant inverse tangent operator, and $\text{wrap}(\cdot)$ normalizes the angular difference to $[-\pi, \pi]$.
	\begin{equation}
	J_{goal}(\tau) = \left| \text{atan2}(y_{g} - y_{end}, x_{g} - x_{end}) - \theta_{end} \right|
	\end{equation} and $J_{obs}(\tau)$ represents the physical obstacle penalty. It is derived from the minimum Euclidean distance $d_{phy}$, between the robot's position $\mathbf{p}(t) = [x(t), y(t)]^T$ along the trajectory and the filtered point cloud generated from the RGB-D camera's depth frame. To ensure collision-free navigation, the cost scales exponentially with proximity,
	\begin{equation}
	J_{obs}(\tau) = \exp\left(-\alpha \cdot \min_{t \in \tau} d_{phy}(\mathbf{p}(t))\right)
	\end{equation} and $J_{semantic}(\tau)$ acts as a virtual obstacle penalty. The visual perception module converts restricted zones (e.g., non-traversable semantic labels) into virtual point clouds with a height of $0.5$m. The planner treats these exactly as physical barriers, applying an identical penalty structure based on the distance $d_{sem}$ to the nearest virtual obstacle,
	\begin{equation}
	J_{semantic}(\tau) = \exp\left(-\beta \cdot \min_{t \in \tau} d_{sem}(\mathbf{p}(t))\right)
	\end{equation}
where $w_g, w_o$, and $w_s$ are weighting constants, and $\ alpha$, $ \beta$ are decay constants. These $\ alpha$ and $ \beta$ are tuned to the repulsiveness of geometric and semantic barriers, respectively. All these parameters are fixed across experiments and empirically tuned. Finally, the candidate trajectory $\tau^*$ minimizing the aggregate cost is selected. The control inputs from the selected optimal trajectory, the linear ($v$) and angular ($\omega$) velocities respectively, are extracted and published to the robot's low-level controller for imediate action on task instruction.
\begin{figure}[t]
	\centering
	\includegraphics[width=1.0\linewidth, trim={6.0cm 5.0cm 6.0cm 3.5cm}, clip]{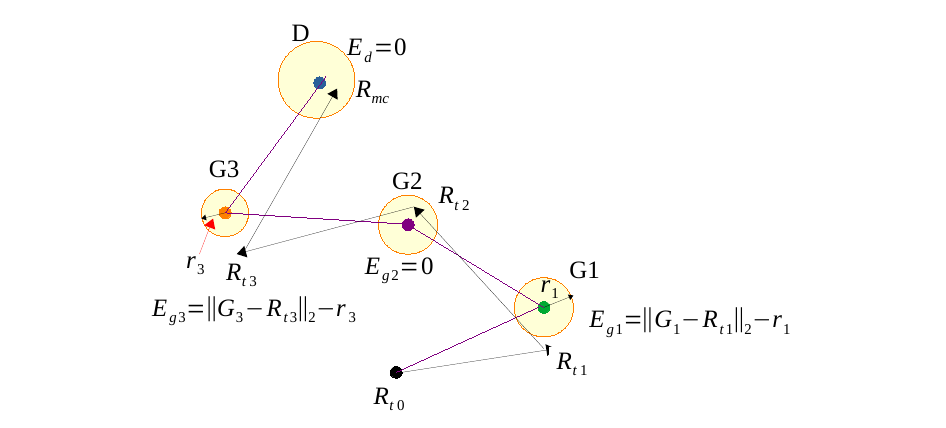}
	\caption{Illustration of odometry drift estimation during navigation. The robot moves through intermediate subgoals $G_1$–$G_3$ toward the final destination $D$, and the drift is computed based on the deviation between the robot pose and the corresponding target.}
	\label{odom}
\end{figure}
Fig.~\ref{odom} shows black arrows that denote the robot’s traversed path between the robot and its next target. The intermediate targets are $G_1$, $G_2$, and $G_3$, while $D$ denotes the user-specified final destination. The light-yellow circular regions shown in Fig.~\ref{odom} represent safety zones centered around each target, whose radii are set by the user based on surrounding visual and geometric context. The odometry drift at each subgoal, denoted as $E_{g1}$, $E_{g2}$, $E_{g3}$, and $E_d$, is computed during navigation, with the final drift evaluated at the destination point $D$. In our work, the navigation is performed without maintaining a global map, localization relies solely on wheel encoder and IMU measurements, and no loop closure is applied.
\section{EXPERIMENTS}
To evaluate the robustness and real-time performance of our framework, validation was conducted in both a high-fidelity simulation environment and a real-world experimental setup. Consistent with the distributed architecture described previously, both setups use a Client-Server architecture to balance computational load efficiently.

In the Simulation setup, the Client node serves as the robot controller and runs on a desktop workstation equipped with an Intel Core i7-4770 processor. The Server node, responsible for heavy perception tasks, runs on a separate high-performance workstation equipped with an NVIDIA RTX A4000 GPU.

\begin{figure}[t]
	\centering
	\includegraphics[width=1.0\linewidth, trim={1.0cm 4.0cm 4.0cm 2.5cm}, clip]{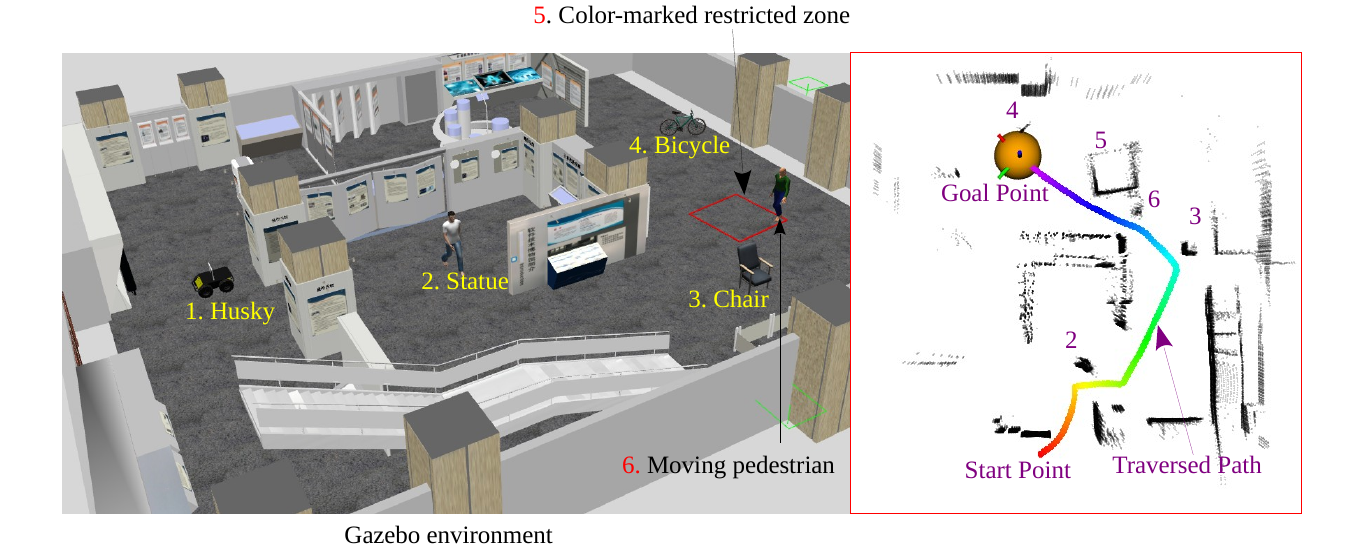}
	\caption{Experimental validation in the Gazebo simulation environment. (Left) The 3D test workspace containing the Husky A200 robot and various semantic obstacles labeled as: 2) Statue, 3) Chair, 4) Bicycle, 5) Color-marked restricted zone, and 6) Moving pedestrian. (Right) Top-down view of the generated point cloud map illustrating the autonomous navigation performance, showing the Start Point, Goal Point, and the resulting collision-free Traversed Path}
	\label{sresult1}
\end{figure}

\textbf{Simulation Results and Analysis:} The proposed framework successfully guided the differential-wheeled robot to its final destination (4. Bicycle) within the Gazebo simulation environment, demonstrating robust autonomous navigation, following the complex multi-step instructions. The robot sequentially localized and approached each semantic landmark. As illustrated in Fig.~\ref{sresult1}, the traversed path is optimal and collision-free. Crucially, the system demonstrated advanced context-aware perception: it strictly avoided the color-marked restricted zone, treating it as a critical safety constraint, and dynamically navigated around a moving pedestrian without interrupting the mission. These experimental results validate the framework's ability to interpret natural language commands while prioritizing environmental safety and efficiency.

\begin{figure}[t]
	\centering
	\includegraphics[width=1.0\linewidth, trim={5.0cm 4.0cm 5.0cm 3.5cm}, clip]{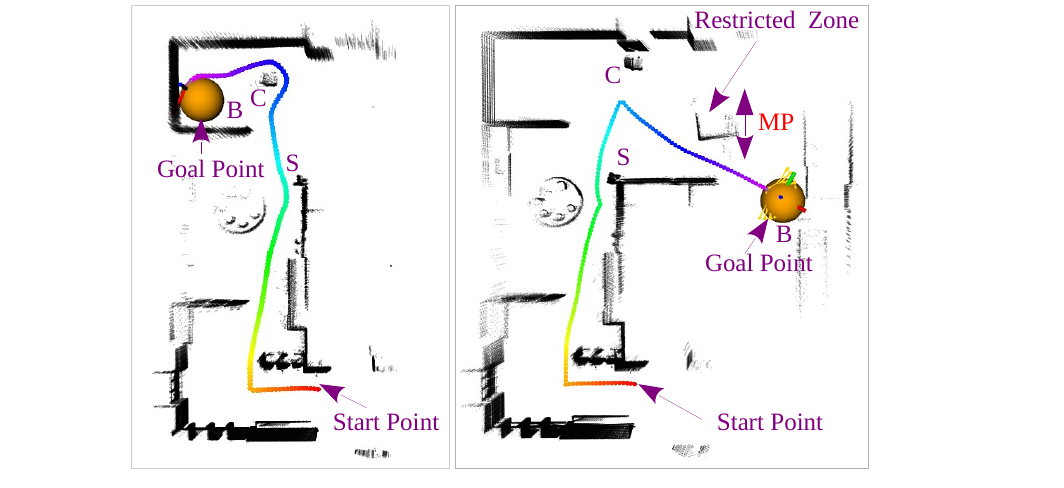}
	\caption{Explored environment and traversed paths for two distinct navigation tasks. The left and right panels display the top-down occupancy map, where black regions represent physical barriers and context-aware constraints. The robot successfully navigates from the Start Point to the Goal Point by executing a sequence of semantic actions. Key landmarks are labeled as S (Statue/Human), C (Visitor Chair), and B (Bicycle). The planner also accounts for dynamic and static constraints, such as the Moving Pedestrian (MP) and the designated Restricted Zone.}
	\label{sresult2}
\end{figure}
\begin{table*}[!t]
	\centering
\caption{Comparison of Modern Navigation Paradigms (Extended). 
	\footnotesize{\cmark~Supported, \xmark~Not supported, $\triangle$~Partial.}}
	\label{tab:method_comparison_extended}
	\resizebox{2.0\columnwidth}{!}{%
		\begin{tabular}{lcccccccc}
			\toprule
			\textbf{Method} &
			\textbf{Global Map} &
			\textbf{Drift Sensitivity} &
			\textbf{Long-Term Nav.} &
			\textbf{Unknown Env.} &
			\textbf{Semantic Understanding} &
			\textbf{Context Awareness} &
			\textbf{User Interaction} &
			\textbf{Replanning} \\
			\midrule
			A* / RRT*        & \cmark & \cmark & \xmark & \xmark & \xmark & \xmark & \xmark & Limited \\
			DiPPer           & \cmark & \cmark & $\triangle$ & \xmark & \xmark & \xmark & \xmark & Limited \\
			ViT-A*           & \cmark & \cmark & $\triangle$ & $\triangle$ & $\triangle$ & \xmark & \xmark & Limited \\
			Vision-Aided A*~\cite{kumar2025vision}   & \cmark & \cmark & \xmark & \cmark & \xmark & \cmark & \xmark & Online \\
			SemExp (Semantic Exploration) \cite{chaplot2020object} & \xmark & $\triangle$ & $\triangle$ & \cmark & \cmark & \cmark & \xmark & Moderate \\
			CARe (Context-Aware Replanning)~\cite{ko2024contextaware} & \xmark & $\triangle$ & \cmark & \cmark & \cmark & \cmark & \xmark & Advanced \\
			SEMNAV (Semantic Segmentation)~\cite{flor2025semnav} & \xmark & \xmark & \cmark & \cmark & \cmark & \cmark & \xmark & Advanced \\
			Multi-Agent Semantic Nav~\cite{liu2022multi} & \xmark & \xmark & \cmark & \cmark & \cmark & \cmark & \xmark & Advanced \\
			\textbf{Ours}    & \xmark & \textbf{\xmark} & \textbf{\cmark} & \textbf{\cmark} & \textbf{\cmark} & \textbf{\cmark} & \textbf{\cmark} & \textbf{Online} \\
			\bottomrule
	\end{tabular}}

\end{table*}
\begin{table}[h]
	\centering
	\caption{Qualitative Evaluation of LLM-Guided Instruction Parsing.}
	\label{tab:llm_parsing_safe}
	\resizebox{\columnwidth}{!}{%
		\begin{tabular}{|p{\columnwidth}|}
			\hline
			\textbf{Case 1: Complex Navigation} \\
			\hline
			\textbf{User Input:} \textit{``Buddy go to the statue (human), take a right turn and move a little ahead, then find the visitor chair to reach close to it, and on the right you may see a bicycle."} \\
			\hline  
			\textbf{LLM Output:} \newline
			\texttt{[\{"action": "FIND", "target": "statue (human)", "id": 0\},} \newline
			\texttt{\{"action": "MOVE", "desc": "turn right...", "f\_dist": 2.5, "yaw": -1.57\},} \newline
			\texttt{\{"action": "FIND", "target": "visitor chair", "id": 56\},} \newline
			\texttt{\{"action": "FIND", "target": "bicycle", "id": 1\}]} \\
			\textbf{Status:} Success \\ 
			\hline
			
			\textbf{Case 2: Metric Constraints} \\
			\hline
			\textbf{User Input:} \textit{``Buddy take left turn, move 4 m ahead, then find the standing person to reach close to it, In your right you may see the visitor chair follow it. Finally go at spot where bicycle is standing."} \\
			\hline 
			\textbf{LLM Output:} \newline
			\texttt{[\{"action": "MOVE", "desc": "left turn 4m", "f\_dist": 4.0, "yaw": 1.57\},} \newline
			\texttt{\{"action": "FIND", "target": "standing person", "id": 0\},} \newline
			\texttt{\{"action": "FIND", "target": "visitor chair", "id": 56\},} \newline
			\texttt{\{"action": "FIND", "target": "bicycle", "id": 1\}]} \\
			\textbf{Status:} Success \\ 
			\hline
		\end{tabular}%
	}
\end{table}
Odometry data is reset after each successful mission completion, limiting drift to the distance between the start point and its target. The maximum subgoal distance is 10 m due to depth data quality. Geometric targets may introduce additional drift for navigation in a featureless Corridor. RGB-D data do not constrain them. Heavy computational SLAM is not required for short-range navigation, but it can be integrated for accurate, precision-critical tasks.The proposed framework does not rely on a persistent global SLAM map. It is a mapless navigation algorithm that enables a wheeled robot to operate solely from real-time camera data and local visual cues. A temporary map frame is introduced for qualitative observation and is fully decoupled from the navigation process. It serves as a fixed reference to depict the vehicle’s trajectory in real time, without contributing any spatial information for path planning or goal execution.

\textbf{Real-World Experimental Results and Analysis:} The real-world experiments are conducted on the Husky robot, which is powered by an onboard NVIDIA Jetson AGX embedded module. The server computing device (optional) is used when the robot system is insufficient for the real-time execution of the semantic visual task. 

\begin{figure}[t]
	\centering
	\includegraphics[width=1.0\linewidth, trim={1.0cm 2.5cm 1.0cm 1.5cm}, clip]{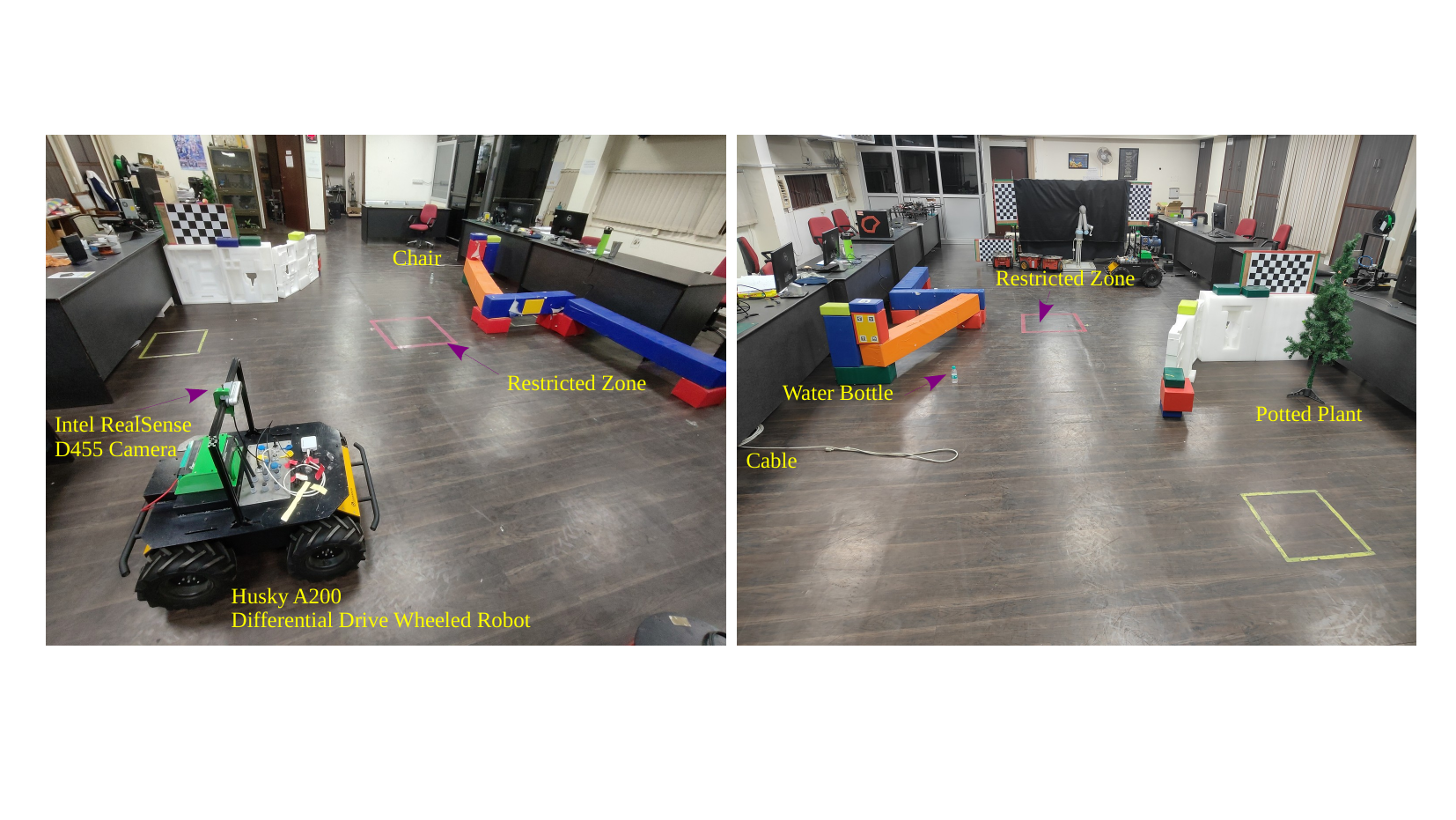}
	\caption{Indoor experimental environment with the Husky A200 robot and onboard Intel RealSense D455 camera. The scene includes multiple visual targets and restricted regions in a confined workspace used for real-world navigation evaluation.}
	\label{rwsetup}
\end{figure}
\begin{figure}[t]
	\centering
	\includegraphics[width=1.0\linewidth, trim={1.0cm 3.0cm 1.0cm 3.0cm}, clip]{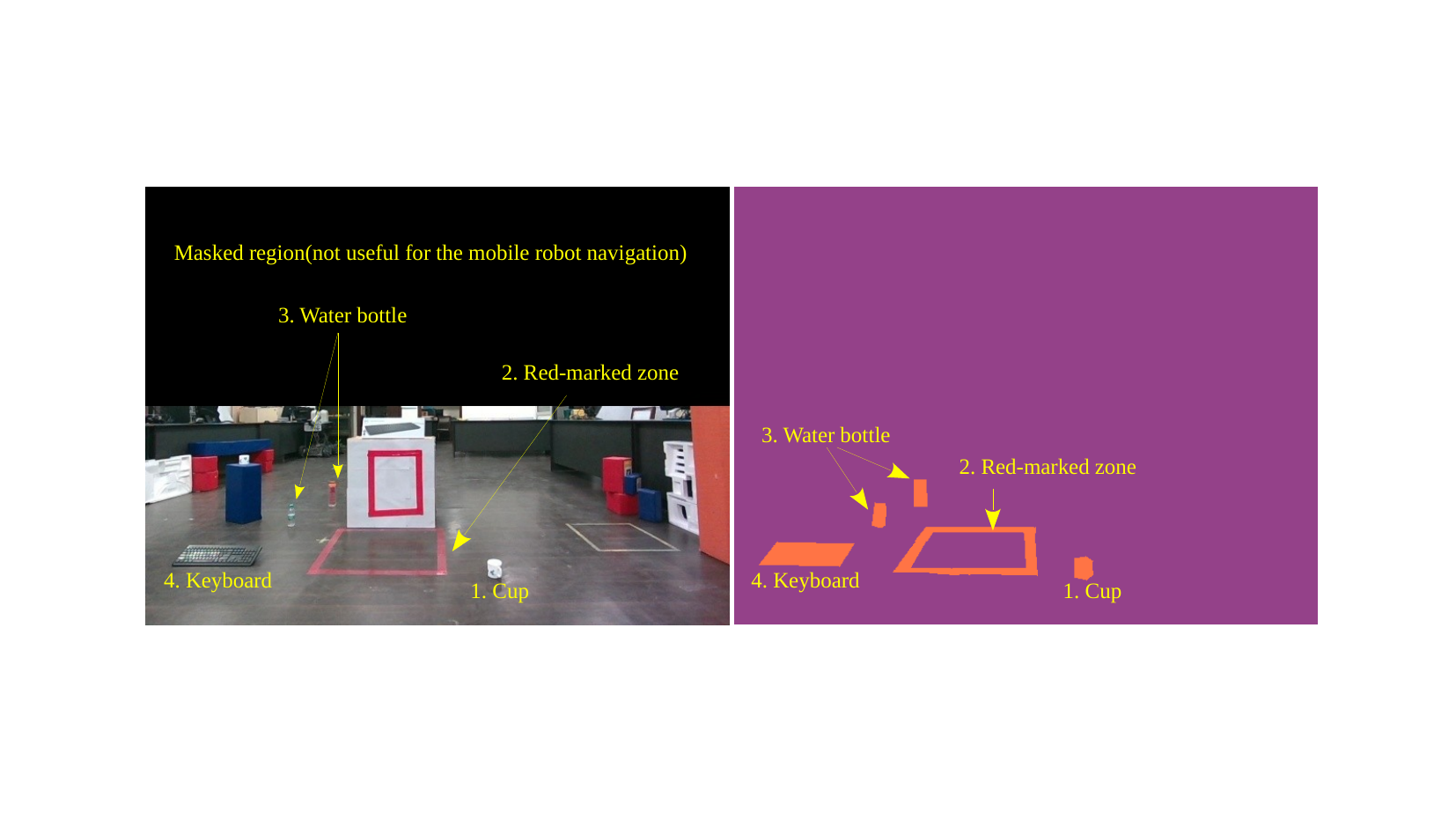}
	\caption{ESANet model input and output: The frame (left) shows the settings: the marked objects are placed on the surface, the red-marked box is drawn on the ground, and the remaining items are at a significant height. The segmentation model output is shown (right), where we show the segmented region of the labeled items, which the user defines.}
	\label{esanet}
\end{figure}
This section evaluates the system's performance in a visually rich, constrained real-world environment ($10\,\text{m} \times 5.5\,\text{m}$), as illustrated in Fig.~\ref{rwsetup}. The environment is cluttered with common physical barriers and objects, including water bottles, a red-marked box, cables, chairs, and potted plants. This indoor setting serves as a rigorous testbed for multiple experimental trials and system validation.
\begin{figure}[!t]
	\centering
	\includegraphics[width=1.0\linewidth, trim={1.0cm 2.5cm 1.0cm 1.5cm}, clip]{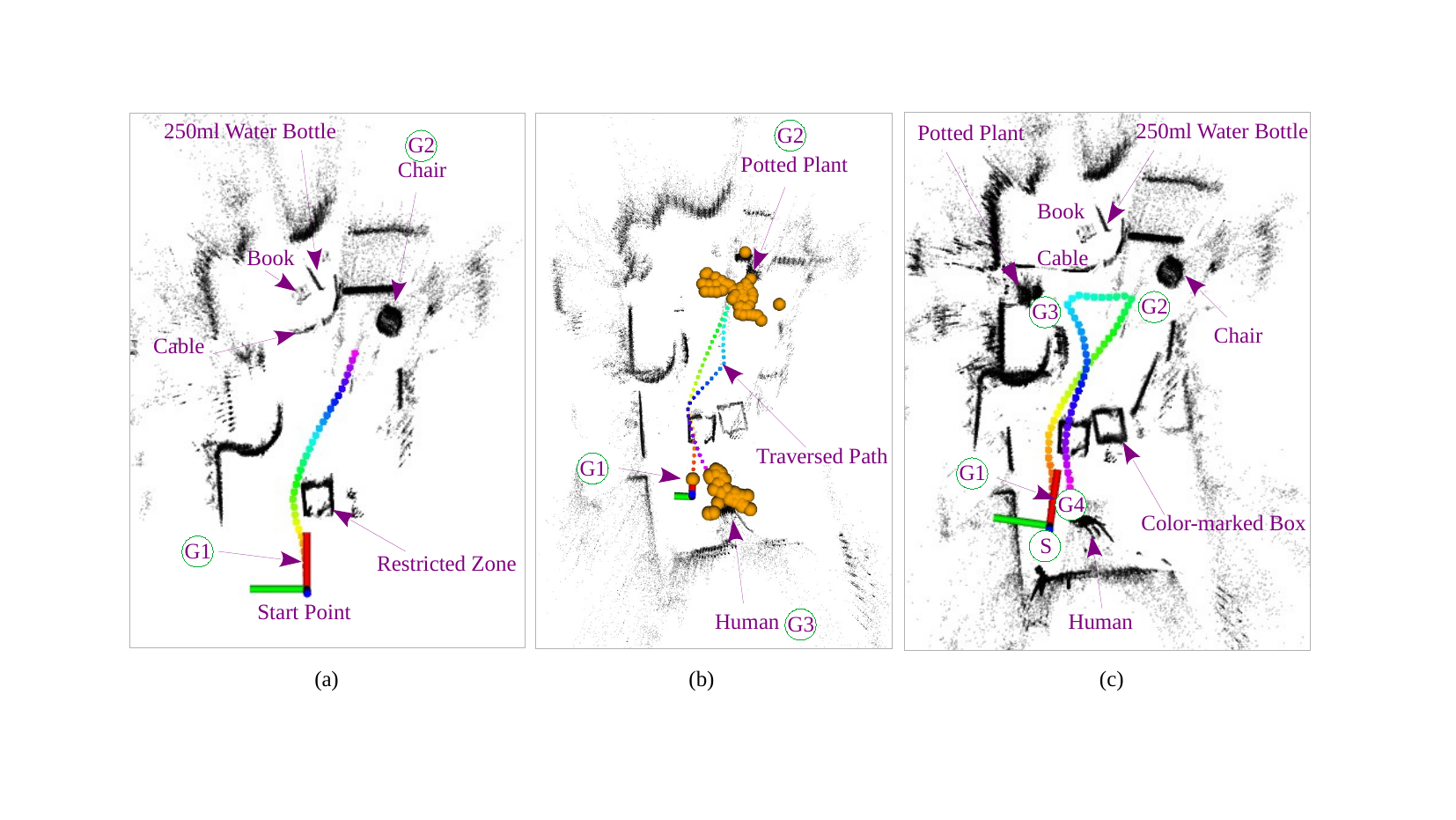}
	\caption{Real-world navigation results across multiple trials. The robot starts from S and first moves toward an egocentric geometric goal (G1), followed by visually detected semantic targets (G2–G4). White regions indicate free space, black regions represent obstacles, and the colored path shows the executed trajectory.}
	\label{rwr1}
\end{figure}
\begin{figure*}[!t]
	\centering
	\includegraphics[width=0.7\linewidth, trim={1.0cm 2.0cm 1.0cm 1.7cm}, clip]{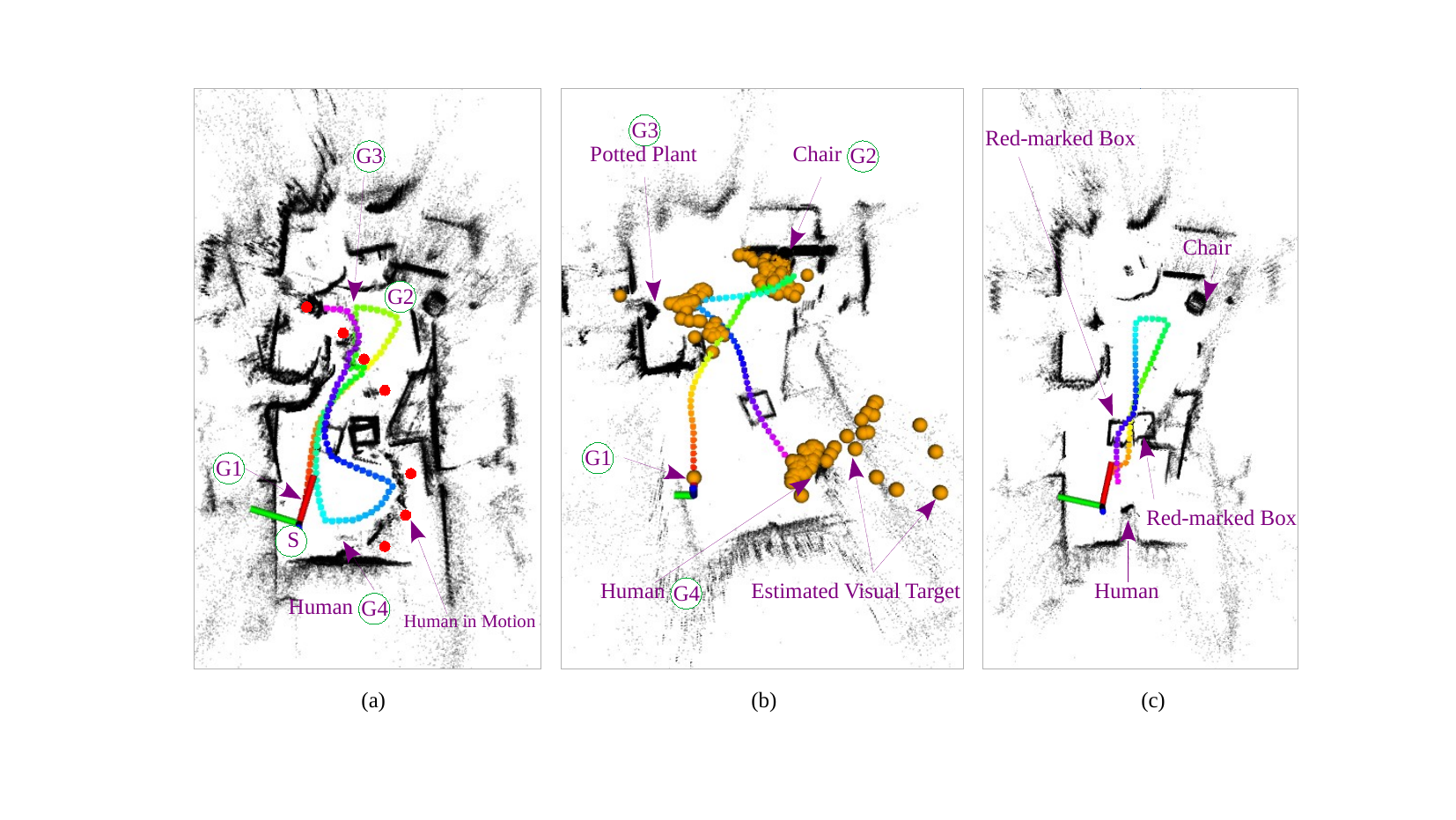}
	\caption{Experimental results of the visual navigation system. (a) The wheeled robot successfully tracking and following a dynamic human target. (b) Estimation of the fixed visual target point; note that despite the spatial spread of the estimated points, the robot successfully navigated to the visual destination. (c) The view of the red-marked box target, showing lateral shifts caused by the robot's to-and-fro motion during navigation.}
	\label{rwr3}
\end{figure*}

Fig.~\ref{rwr1}~(a) demonstrates the robot's ability to navigate context-aware settings. The robot successfully identifies and avoids the red-marked box (prohibited zone), then moves toward the visual target (a chair) and finally halts at a definite distance threshold. The top-down point cloud view in Fig.~\ref{rwr1}~(a) reveals the detection of smaller objects such as a Book, a Cable, and a 250ml bottle. Notably, our proposed method effectively filters semantic noise located on the ground surface near the target chair, ensuring robust target identification.

The efficacy of the proposed approach in sequential navigation is demonstrated in Fig.~\ref{rwr1}~(b). The robot takes a user task-instruction to navigate and reach the egocentric geometric subgoals, then the semantic subgoals ($G_2$ and $G_3$). During this trajectory, the system detects a double red-marked box and visualizes multiple golden spheres, which represent estimated visual targets generated during navigation prior to reaching the current goal. It was observed that semantic segmentation noise, caused by varying lighting conditions and reflective surfaces, occasionally introduced false positive points into the environmental point cloud; however, the navigation system remained robust to these disturbances.

Finally, Fig.~\ref {rwr1}~(c) presents the results of a complex navigation task involving four distinct local goals. A minor map layout shift is observed in the global map; despite this slight deviation, the robot completed the multi-goal mission, confirming the system's robustness in complex, multi-goal navigation tasks.
\begin{table}[t]
	\centering
	\caption{Inference Time Performance Across Computing Platforms}
	\label{tab:inference_time}
	\resizebox{\columnwidth}{!}{%
		\begin{tabular}{@{}llcl@{}}
			\toprule
			\textbf{Module} & \textbf{Task} & \textbf{Latency (ms)} & \textbf{Execution Mode} \\ 
			\midrule
			
			\multicolumn{4}{l}{\textbf{Server Platform (NVIDIA RTX A4000)}} \\ 
			\midrule
			RedNet           & RGB-D Segmentation  & $29.2 \pm 1.8$       & Continuous \\
			ESANet           & RGB-D Segmentation  & \textbf{$34.8 \pm 0.7$} & Continuous \\
			OSNet            & Person Re-ID        & $50 \pm 5$           & On-demand \\
			YOLOv8           & Object Detection    & $12.3 \pm 1.4$       & On-demand \\
			LLM (Gemini Pro) & Instruction Parsing & $150 \pm 30$         & Per command \\
			
			\midrule
			\multicolumn{4}{l}{\textbf{Onboard Platform (Husky A200 – Jetson AGX Xavier)}} \\ 
			\midrule
			RedNet           & RGB-D Segmentation  & $87 \pm 7$           & Continuous \\
			ESANet           & RGB-D Segmentation  & $108 \pm 14$         & Continuous \\
			OSNet            & Person Re-ID        & $80 \pm 10$          & On-demand \\
			YOLOv8           & Object Detection    & $38 \pm 8$           & On-demand \\
			LLM (Gemini Pro) & Instruction Parsing & $190 \pm 35$         & Per command \\
			
			\bottomrule
	\end{tabular}}
\end{table}

The proposed framework is validated in a dynamic scenario where the robot follows a moving human target, as depicted in Fig.~\ref{rwr3}~(a). Throughout this task, the planner successfully generates trajectories that avoid collisions and strictly adhere to user-defined environmental constraints, specifically avoiding the red-marked box and cables.

Fig.~\ref{rwr3}~(b) highlights challenges in narrow passages, where the path overlaps with the prohibited zone due to non-holonomic constraints and limited turning radius, further affected by the robot’s large footprint~(988\, mm $\times$ 670\, mm) and skid-steer kinematics; however, the framework remains platform-agnostic and can be readily deployed on smaller mobile robots. The figure also shows estimated target points (golden spheres), and despite their high variance, the algorithm effectively filters them to produce a safe, converging path.

The impact of sensor noise is analyzed in Fig.~\ref {rwr3}~(c). Here, artifacts arising from semantic segmentation-to-point cloud projection introduced geometric errors. Consequently, the robot momentarily grazed the red-marked zone during the initial motion from the start pose to the visual target (chair). However, the tracking from the potted plant to the human showed that the robot corrected and safely avoided the no-go area.\\
\begin{table}[t]
	\centering
	\caption{Performance Comparison of Navigation Modalities Under Different Environmental Constraints}
	\label{tab:modality_comparison}
	\resizebox{\columnwidth}{!}{%
		\begin{tabular}{l c c c c}
			\toprule
			\textbf{Scenario} & \textbf{Constraint} & \textbf{Geometric} & \textbf{Visual} & \textbf{Hybrid (Ours)} \\
			\midrule
			Featureless Corridor & No visual landmarks & 92.0 & 15.0 & \textbf{95.0} \\
			Symmetric Room & Geometric ambiguity & 40.0 & 94.0 & \textbf{96.0} \\
			Occluded Target & Target not visible & 85.0 & 30.0 & \textbf{91.0} \\
			Precise Positioning & Exact stopping required & 96.0 & 65.0 & \textbf{98.0} \\
			Complex Mission & Sequential task execution & 10.0 & 25.0 & \textbf{88.0} \\
			\bottomrule
		\end{tabular}
	}
\end{table}
\begin{table}[t]
	\centering
	\caption{Performance Evaluation Results Over 50 Trials per Scenario. The simulation scenarios (Indoor, Garage, Campus, Forest, Tunnel) are given in detail~\cite {cao2022autonomous, yang2022far}. These setups are modified to include visual objects and semantic context-aware items (soft collision) for our system performance evaluation. We consider the environmental structure (hard collision) that obstructs the navigation path.}
	\label{tableFive}
	\resizebox{\columnwidth}{!}{%
		\begin{tabular}{@{}lccccc@{}}
			\toprule
			\textbf{Environment} &
			\textbf{Success Rate (\%)} &
			\textbf{Soft Collision (\%)} &
			\textbf{Hard Collision (\%)} &
			\textbf{Avg. Path Length (m)} &
			\textbf{Drift Error (m)} \\
			\midrule
			\multicolumn{6}{@{}l@{}}{\textit{Simulation}} \\
			Indoor  & 100 & 10.0 & 0.0 & 12.4 & $0.03 \pm 0.01$ \\
			Garage  & 100 & 8.5 & 0.0 & 20.8  & $0.05 \pm 0.03$ \\
			Campus  & 98 & 10.0 & 0.0 & 30.5  & $0.05 \pm 0.02$ \\
			Forest  & 95 & 15.2 & 0.3 & 15.5  & $0.08 \pm 0.01$ \\
			Tunnel  & 98 & 14.0 & 0.0 & 25.2  & $0.05 \pm 0.02$ \\
			\midrule
			\multicolumn{6}{@{}l@{}}{\textit{Real-World Experiments (Husky A200 Platform)}} \\
			Indoor Lab (Fig.~\ref{rwsetup}) & 95 & 15.0 & 0.5 & 12.4 & $0.1 \pm 0.02$ \\
			Outdoor Area & 98 & 5.6 & 0.0 & 124.6 & $0.05 \pm 0.01$ \\
			\bottomrule
	\end{tabular}}
\end{table}
The average odometry drift error reported in Table~\ref{tableFive} is computed as
\begin{equation}
\bar{E}_{\text{drift}} =
\frac{1}{N}
\sum_{t=1}^{N}
\max\left(0,\;
\left\| \mathbf{G}_d^{(t)} - \mathbf{R}_d^{(t)} \right\|_2
- d_s
\right),
\end{equation}
where $\mathbf{R}_d^{(t)}$ denotes the robot's final position at the mission completion time for the $t$-th trial, and $\mathbf{G}_d^{(t)}$ represents the corresponding mission destination location in the global coordinate frame. The parameter $d_s$ defines a safety radius centered at the goal location, within which the robot is considered to have successfully reached the target. This radius is selected based on the physical size of the object and task-specific safety requirements. The variable $N$ denotes the total number of individual missions conducted under identical experimental settings.
 
\section{Conclusion}
In this work, we present a reliable solution for human-language-driven navigation that leverages the synergy between LLM and semantic visual perception of an environment. It empowers service robots to interpret and perform natural language commands, making human-robot interaction more user-friendly and efficient. The experimental results demonstrate that the robot's performance in navigating with its admin intent in the dynamic real-world environment effectively distinguishes between physical barriers (e.g., Geometric objects that may obstruct the navigation path) and soft barriers (e.g., no-go zones and defined critical objects) using a context-aware framework. 
Our future task is to address the unstable environment, lighting conditions, and skidding surfaces handling, and plan to refine the candidate local trajectory generation to better accommodate the robot's kinematic constraints for non-jerking motion.
\bibliographystyle{ieeetr} 
\bibliography{ref} 
\end{document}